\documentclass{math}
\english

\Author{Rastislav Telg\'arsky}

\Title{Multi-target tracking algorithms in 3D}

\ShortTitle{Multi-target tracking algorithms in 3D}

\Abstract{Ladars provide a unique capability for identification of objects and motions in scenes with fixed 3D field of view (FOV). This paper describes algorithms for multi-target tracking in 3D scenes including the preprocessing (mathematical morphology and Parzen windows), labeling of connected components, sorting of targets by selectable attributes (size, length of track, velocity), and handling of target states (acquired, coasting, re-acquired and tracked) in order to assemble the target trajectories. This paper is derived from working algorithms coded in Matlab, which were tested and reviewed by others, and does not speculate about usage of general formulas or frameworks.}


\Address{Central New Mexico Community College}{525 Buena Vista Dr. SE, Albuquerque, NM 87106, U.S.A.}{rtelgarsky@cnm.edu; rastislav@telgarsky.com}

\Subjclass{M10; P50.}  

\begin{document}

\SetUpTitle
\thispagestyle{FirstPage}
\setcounter{page}{165}

\section*{1 Data generated by Ladar}

The photon counting Ladar (laser radar) consists of the illumination laser (transmitter) and the sensor (receiver) at the same location (the monostatic setup). The focus of this paper is on algorithms for processing the images resulting from a prototype of a photon-counting Geiger-mode sensor with an array resolution of 32 $\times$ 32. The sensor has FOV of 50 by 50 meters at 10 km, and a frame rate of 12 kHz. Each frame contains the energy returned from one laser pulse. We construct a 3D image from 200 consecutive frames by calculating the empirical PDF of the range (depth) values; actually, the voxel intensities are defined by the 3D histogram values.

The record of Ladar activity is a file stored on a hard disk. One second of data contains 12,000 2D images, each of which has size 32 $\times$ 32, and each pixel stores 16 bits of data. We use 200 2D images to create one 3D frame. Thus, the frame speed reduces from 12 kHz down to 60 Hz. The file with 300 cumulative frames represents 300 / 60 = 5 minutes of real running time and contains 300 $\times$ 200 = 60,000 of raw 2D images. The size of this file is about 125 MB.

The groups of 200 raw images correspond to one train of 200 pulses of the laser; each of the 32 $\times$ 32 images is the return from a single laser pulse. The pixel values in this image either correspond to a relative time between pulse emission by the laser and the pulse detection by this sensor pixel, or they represent the ceiling (when we stop waiting for return).

The integration time for one 32 $\times$ 32 image is 100 microseconds. If during this time interval of 100 microseconds a pixel is not hit by a photon, then its value is set to the ceiling value. We interpret the non-ceiling values of pixels as the range bins.

Due to very weak signal (laser is eye-safe), and thus very small number of photons per each 32 $\times$ 32 image (resulting from one pulse), we get very little information about the scene. Moreover, the pattern of each image is heavily distorted by a noise which is generated mostly by the sensor.

Each 3D image is created from 200 32$\times$32 images as follows: for each pixel $(x,y)$ we calculate the histogram of occurrences of range bins between an offset (say, 10) and ceiling $-$ offset. Now, $g(x,y,z)$ is the count of the photons detected by the sensor at $(x,y)$ and reflected from the distance $z$. There are no other transmitters in the area having the same frequency as the illuminating laser. It turns out that this new 3D image has size 32 $\times$ 32 $\times$ 600.

\section*{2 Processing of 3D images}

The raw 3D image is populated with too many non-zero voxels, some of which represent the detection noise. Since voxels representing noise have low values (typically, 1 or 2 photons), we could remove the noise by thresholding. However, this will not remove many of the outliers which correspond to small objects in the scene. There is a delicate balance between removing outliers and losing the shapes of objects of interest.\smallskip

\noindent The following noise-reducing functions are user-selectable:\smallskip
\begin{enumerate}
  \item Thresholding
  \item Thresholding and majority rule
  \item Parzen Windows and thresholding
\end{enumerate}
\smallskip

The selectable thresholding algorithms are: a fixed threshold, a fixed percentage of peak voxel value, a fixed convex combination of fixed percentage of peak voxel value and the past value (the moving average), or another hybrid threshold.

The majority rule is taken from mathematical morphology, which uses local processing to establish new image values. One of the schemes is the following. We consider an internal voxel and its neighborhood 3 $\times$ 3 $\times$ 3. We consider the number of nonzero voxels in this neighborhood. This number is between 0 and 27. We make the following decision rule. We set the new value of the initial voxel to 1 or 0 depending on whether the number of nonzero voxels is greater than 2 or not, respectively. Actually, the majority rule (according to the original definition) operates on binary images, so we use first the thresholding to indicate whether the voxel value is significant or negligible, and then we use the majority rule. The majority rule is a voting scheme applied to voxel neighborhoods in order to set their new voxel values.

The Parzen Windows Probability Density Function Estimation, or Parzen Windows - for short, is an approach for an estimation of the "true" probability density function by approximating the local density points. One way of applying Parzen Windows is to take the convolution of $g(x,y,z)$ with a suitable 3D Gaussian density function. This density function depends on three scale parameters: the standard deviations in each of three dimensions. It seems from examples processed so far that Parzen Windows provide more details of target shape while also remove noise from the scene.

\section*{3 Connectivity and labeling}

The processing creates a binary mask for the 3D image indicating the potential target locations. Now we find the connected components in the scene with 3D morphological connectivity operators available in Matlab. The structural elements can be selected with 6, 18, or 26 neighbors when based on 3 $\times$ 3 $\times$ 3 neighborhood. We declare the list of these connected components as the acquired targets. Targets are numbered from 1 to $N$, and k-th target consists of voxels with value $k$. If $N$ is greater than the selected maximal number of targets tracked, $T_{max}$, then we reduce the list as follows. We sort the list by means of an importance criterion in the decreasing order, so that the most important target appears as the first on the sorted list. Then we truncate the list to the length $T_{max}$. The importance criterion can be a combination of the following: target volume (number of voxels), speed, direction of motion, variance of motion, etc.

\section*{4 Target associations}

After the first step in acquisition we have only $T_1 \leq T_{max}$ of newly acquired targets. However, in step n $>$ 1, we have $T_{n-1} \leq T_{max}$ targets from previous step $n-1$ and also $T_n \leq T_{max}$ of the newly acquired targets at this step n. Both lists are sorted by decreasing importance criterion. We must find the best matches between the 'old'\ targets and 'new'\ targets, that is, which instances of targets in the current step n are extensions of target in the accumulated list indexed with $T_{n-1}$. If the association criteria are too strong, we may end up with no association, and therefore we must accept $T_{n-1} + T_n$ targets for further processing, in particular, for sorting by the importance criterion, and then limiting their total number by $T_{max}$.

\begin{figure} [h!]
\begin{center}
\includegraphics[width=10cm]{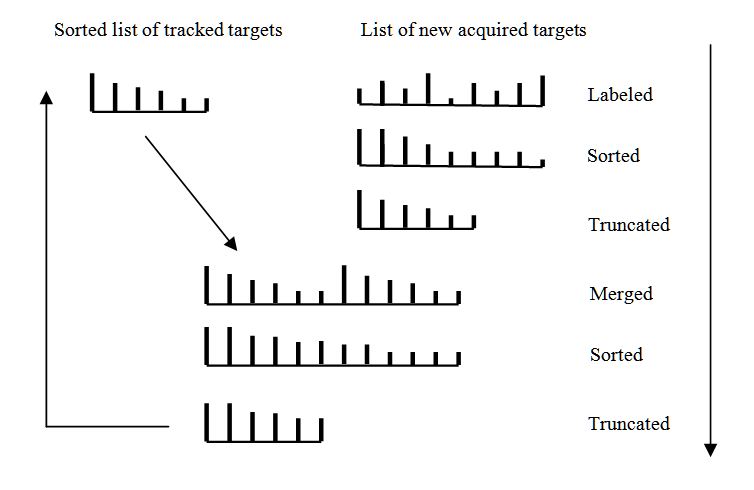}
\end{center}
\begin{center}
Figure 1: Merging of target lists
\end{center}
\end{figure}

There are several methods of associating (matching) targets indexed by $T_{n-1}$ to targets indexed by $T_n$. Here we form the target association matrix $M$ with $T_{n-1}$ rows and $T_n$ columns, and define each entry as the numerical value of the relationship strength between an 'old'\ and a 'new'\ target. In the extreme, each 'old'\ target would be related to (associated with) a 'new'\ target, so all entries of the matrix $M$ would be positive numbers. This would lead to an unmanageable complexity of the tracking program. Each entry of the matrix $M$ can be considered as a weighted combination of correlations between attributes of 'old'\ target and attributes of 'new'\ target.

As we sort the target lists by a single criterion or a combination of criteria, we generate the entries of the association matrix $M$. Since the most undesirable feature in target tracking is a target jumping from a location to a location violating the continuity of motion, we use the target location as the main criterion for association. However, there are different ways to characterize the target location. The most popular is the target centroid, but we use here the target location characterized by the bounding box, which is also used in many popular program such as AutoCAD. Thus the target location is characterized by 6 numbers: minima and maxima of $x,y,z$ coordinates of the target voxels.

We define a deterministic matching (association) criterion via bounding boxes. We shall say that an 'old'\ target matches a 'new'\ target if the bounding box of the 'old'\ target is contained in an expansion of the bounding box of a 'new'\ target, and vice versa. The expansion of a bounding box is defined by decreasing the minima and increasing the maxima in each dimension by a fixed number (positive integer). This number is specified before running the tracking program.

We say that the 'new'\ target is a valid extension of an 'old'\ target, if the above symmetric relation holds. The next refinement of this notion is using the Kalman filter to predict the six values from the 'old'\ target’s bounding boxes and try to match the bounding box of 'new'\ target to these predicted values. In this case each 'old'\ target must have its own Kalman filter stored in the history array. As an alternative, we use the Kalman filter predictor for centroid, so we match a 'new'\ target to an 'old'\ target if its centroid is close enough to the predicted value of the centroid of the 'old'\ target (from step $n-1$ to the step $n$).

\section*{5 Target states}

After performing the matching procedure we end up with targets in 4 different states. To 'old'\ targets which were not matched to 'new'\ targets we give some chances for a short period of time, typically for 3 time frames; this number is selectable from the menu. This chance to be detected and matched later but not too late is called coasting. Here is the description of target states.

\textit{State 1 (yellow).} It is the coasting state. The target was either detected in step $n-1$ but not matched (so, it is going into coasting state with bad count = 1) or target was in coasting state in step $n-1$, it is not matched to any 'new'\ target in the current step $n$, and the number of bad counts in the consecutive coasting states did not exceed 3. If the number of consecutive coasting states is 4 and the target is not matched to any 'new'\ target, then the target is not recorded in the tracking history array. We say, that after 3 unsuccessful trials, the target is declared to be lost.

\textit{State 2 (green).} The target was detected (in the current step $n$), but it was not matched to any target from the previous step $n-1$. So, it is considered to be a 'new'\ target.

\textit{State 3 (red).} The target was detected in the current step n and matched to a coasting target in the step $n-1$, while the count of consecutive coasting states did not exceed 3. In another words, the previous target state was 1 (yellow) and the number of these coasting states did not exceed 3. We say that this target was re-acquired.

\textit{State 4 (magenta).} The target is detected in the current step $n$ and is matched to a valid (non-coasting) target in step $n-1$.

The target states are recorded in the tracking history array. This array is circular, maintained in each tracking step, and its length is 10. This number is larger than 7, which is the maximal selectable number of coasting states.

The total number of these 4 kinds of targets does not exceed $2 \times T_{max}$. But, if it is greater than $T_{max}$, we shall sort the target list again by the same criteria as was sorted the list of newly detected (and labeled) targets. Note that the optimal sort here is the merge sort, because both lists are already sorted according to the same criterion. Then we truncate the list to $T_{max}$, thus keeping the most important targets only. This operation, however, is pretty drastic, because we are removing the relationships (matches, associations) of the removed targets to the remaining targets. Of course, we must work with the list of length $2 \times T_{max}$ in order to sort it and then reduce it to the length $T_{max}$.

\section*{6 Target trajectories}

We developed an alternative scheme to the association matrix $M$. Our scheme consists of two arrays which record forward and backward links of targets. Both of them extend target trajectories by one step if a match happens at time $n$. First, we should note that the matching relationship does not have to be symmetric. If the matching relation is not symmetric, then we end up with two association matrices $M_1$ and $M_2$, where $M_1$ records the matching of an 'old'\ target to a 'new'\ one and $M_2$ records the matching of a 'new'\ target to an 'old'\ one. We use the index-valued matrices, FWlink and BWlink, to store the match relations. Along with these deterministic relationships we could also store additional values assigned to these relationships, but it would require to define and maintain two more arrays. The relation
$$
FWlink_{n-1,mt} = nt
$$
means that the target with index $mt$ in step $n-1$ (an 'old'\ target) is linked forward to step $n$ with the 'new'\ target having the index $nt$. Similarly, the relation
$$
BWlink_{n, nt} = mt
$$
means that the target having index nt at the current step $n$ is linked backward to the target with the index mt at the step $n-1$.

\begin{figure} [h!]
\begin{center}
\includegraphics[width=8cm]{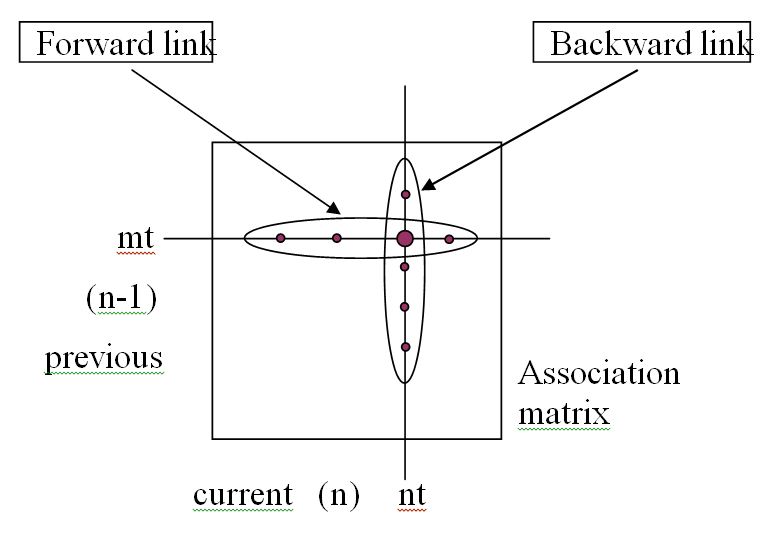}
\end{center}
\begin{center}
Figure 2: Target associations
\end{center}
\end{figure}

Of course, recording one type of link is sufficient, especially in the situation when the matching relation is symmetric. However, keeping two links, the forward link and the backward link, allows for two different reconstructions of target trajectories.

\section*{7 Feature extraction, display and output}

The purpose of feature extraction is usually two-fold. Some features are used for tracking the targets and some for target discrimination. For each target we record 23 features. Past targets with their features are remembered up to 10 frames back (in the history array). For matching (association) we could use predictors depending on up to 10 steps back, however, the stationary approach was working well with experimental data provided.

The tracking algorithms can distinguish between moving targets and static targets. Targets can split or merge any time. A target can enter the FOV any time as well as exit it. There are no prior assuptions about photon distributions, target sizes or target speeds. For all targets the important features are: the size (volume, number of voxels), the location, and the orientation. For moving targets we have also speed and acceleration. Those are the instant features. With target we can also associate the trajectories in $x, y$ and $z$, or the time series of other attributes. Finally, each target lasts during entire tracking sequence, or is short-lived (ephemeral), or something between these extremes.

In general, an object can be tracked if it has a location and size, and we are able to estimate them.

We can plot trajectories of each target as three separate plots of $x, y$, and $z$ as functions of time, or as one (parametric) 3D plot. The frames (edges) of boxes are displayed around targets tracked in 3D. At the end of tracking sequence, a video in AVI format can be saved.

\section*{8 Conclusion}
Thousands of publications and internal research reports contain many diverse algorithms for target tracking which depend on initial specifications, hardware available and/or hardware build for the project, and software written for the projects. This paper describes a very special case of multiple target tracking with Ladar, but also general purpose algorithms having mathematical, engineering and educational values.

\section*{Acknowledgement}
My thanks are due to Boeing-SVS engineering teams, especially, to Michael Salisbury, Matthew Heino and Elizabeth Rejman.

\end{document}